\documentclass[conference]{IEEEtran}
\IEEEoverridecommandlockouts

\newif\ifanonymous
\anonymousfalse 

\usepackage{cite}
\usepackage{amsmath,amssymb,amsfonts}
\usepackage{algorithmic}
\usepackage{graphicx}
\usepackage{textcomp}
\usepackage{xcolor}
\usepackage{booktabs}
\usepackage{array}
\usepackage{url}
\usepackage{float}

\def\BibTeX{{\rm B\kern-.05em{\sc i\kern-.025em b}\kern-.08em
    T\kern-.1667em\lower.7ex\hbox{E}\kern-.125emX}}

\begin{document}

\title{RIDE: Relocalization-Informed Depth Estimation with 3D Gaussian Splatting}

\ifanonymous
\else
\author{
\IEEEauthorblockN{Jiarong Lian\IEEEauthorrefmark{1}, Zhe Xiao\IEEEauthorrefmark{2}, Zhaoyang Zhang\IEEEauthorrefmark{1}, Wei Li\IEEEauthorrefmark{1}, and Ruizhi Chen\IEEEauthorrefmark{1}}
\IEEEauthorblockA{\IEEEauthorrefmark{1}The Chinese University of Hong Kong, Shenzhen}
\IEEEauthorblockA{\IEEEauthorrefmark{2}Carnegie Mellon University}
}
\fi

\maketitle

\begin{abstract}
Render--match--PnP relocalization establishes correspondences between query image pixels and 3D map points for camera pose recovery, but their potential to support dense depth estimation is often overlooked. To exploit this geometric information, we present RIDE, which estimates dense metric depth from a robot's RGB stream. Given a metrically scaled 3D Gaussian Splatting (3DGS) model, RIDE combines sparse metric depth observations derived from PnP-RANSAC inlier correspondences with the geometric prior of a pretrained video-depth model. To handle uneven and intermittent observations, it integrates global and local depth correction with temporal memory, supporting depth estimation through short observation gaps after metric scale initialization. Trained on public RGB-D videos, RIDE is evaluated on robot sequences without fine tuning. Experiments show improved depth accuracy and temporal consistency over scale-only calibration, demonstrating how localization geometry can support both pose recovery and dense robot perception.
\end{abstract}


\section{Introduction}
\label{sec:introduction}

Pose estimation and dense metric depth estimation provide complementary information for robot perception. Pose estimation determines the robot's position and orientation in a reference frame, while metric depth describes surrounding 3D geometry for navigation, obstacle avoidance, mapping, and interaction. Despite relying on the same visual input, the two tasks are typically addressed separately.

Render--match--PnP relocalization with a metrically scaled 3DGS model~\cite{kerbl2023gaussian} directly connects these tasks. Given a query image and a coarse pose estimate, the system 1) renders an RGB image and its corresponding metric depth map; 2) matches the rendered and query images; 3) back-projects each matched rendered pixel to a 3D map point; 4) filters the resulting 2D to 3D correspondences with PnP-RANSAC; and 5) recovers the 6-DoF camera pose from the retained geometrically consistent inliers~\cite{liu2025gscpr}. Existing render--match--PnP pipelines typically use the retained correspondences only for pose estimation, leaving the metric information encoded by these correspondences unused.

A straightforward way to obtain dense metric depth for the query view is to render the 3DGS model at the refined pose. However, the resulting depth map may not accurately represent the scene currently observed by the robot: dynamic or previously unseen objects are absent from the map, while incomplete map geometry and rendering artifacts can introduce erroneous depths. In contrast, sparse metric depth observations derived from PnP-RANSAC inlier correspondences are geometrically verified. They are therefore less susceptible than the dense rendered depth map to discrepancies between the map and the current scene.

RGB-guided depth completion can recover dense depth from sparse metric observations, but many approaches operate frame by frame~\cite{ma2018sparsetodense,zhang2023completionformer}. Temporal completion methods also integrate observations across frames~\cite{kim2024rayfusion,zhu2025svdc}. Here, we combine a frozen VDA-S~\cite{chen2025videodepth} prior, computed from a causal image window, with sparse metric anchors derived from PnP inliers. The prior supplies temporally coherent relative scene structure, while the anchors connect it to the map's metric scale.

This combination still requires spatial and temporal calibration. A single global scale cannot fully correct spatially varying errors in the prior, motivating local spatial correction guided by the sparse and unevenly distributed anchors. Moreover, temporal consistency of the relative prior does not ensure stability after metric calibration: changes in anchor count, distribution, and reliability can make independently estimated scales fluctuate. When too few reliable anchors remain, current-frame calibration may become unavailable. These effects motivate global-scale memory to retain and update metric scale, together with temporal correction that aligns and reuses local calibration information from earlier frames. Once metric scale is initialized, this accumulated information can also support depth estimation through short anchor outages.

In this paper, we propose Relocalization-Informed Depth Estimation (RIDE), which estimates dense metric depth from a robot's RGB stream using sparse metric observations derived from PnP-RANSAC inlier correspondences. RIDE consists of three components: a robust global-scale calibration module, a local spatial correction module, and a flow-guided temporal memory module. The global module combines learned anchor reliability with recurrent scale estimation, while the spatial module corrects local errors in a frozen video-depth prior. The temporal module aligns and reuses earlier calibration features to stabilize depth predictions under uneven and intermittent observations.

The main contributions of this work are:
\begin{itemize}
    \item We derive sparse metric depth anchors from PnP-RANSAC inlier correspondences during metric 3DGS relocalization, linking pose recovery with dense depth estimation without additional sensors at deployment.
    \item We introduce RIDE, which combines learned anchor reliability, robust recurrent global-scale calibration, bounded spatial correction, and flow-guided causal memory to calibrate a frozen VDA-S prior despite sparse, unreliable, or missing anchors.
    \item We evaluate RIDE on scene-disjoint public RGB-D sequences and 27 real robot routes across five scenes. It achieves the highest $\delta_1$ and the lowest held-out error and temporal inconsistency on the public benchmark and, among the evaluated methods, the best dense depth estimation for all three real robot metrics.
\end{itemize}

\section{Related Work}
\label{sec:related_work}

\subsection{Visual Relocalization and 3DGS}

Visual relocalization spans several formulations. Absolute pose regression directly predicts a query image's 6-DoF pose in a scene coordinate frame, as in PoseNet~\cite{kendall2015posenet} and MapNet~\cite{brahmbhatt2018mapnet}. Scene coordinate regression predicts 3D scene coordinates at image locations and recovers camera pose through robust geometric estimation, as in DSAC~\cite{brachmann2017dsac} and ACE~\cite{brachmann2023ace}. Feature-matching pipelines instead establish correspondences between a query image and registered reference images, often within a coarse-to-fine retrieval and matching architecture~\cite{sarlin2019coarse}. When reference geometry is available, these image correspondences can be associated with 3D map points for pose estimation. Detector-free matchers such as LoFTR and RoMa provide semi-dense or dense image correspondences without explicit keypoint detection~\cite{sun2021loftr,edstedt2024roma}.

3DGS~\cite{kerbl2023gaussian} has recently emerged as a useful representation for visual localization because it provides an explicit, renderable scene model with efficient differentiable rasterization. GS-CPR renders RGB and depth from a coarse pose, associates image matches with 3D scene points, and refines the pose with PnP-RANSAC~\cite{liu2025gscpr}; GSFeatLoc lifts rendered matches into 2D to 3D correspondences for PnP~\cite{lee2025gsfeatloc}; STDLoc performs sparse-to-dense relocalization with a feature Gaussian representation~\cite{huang2025stdloc}; and Gaussian Splatting Feature Fields align query and rendered features for accurate pose refinement~\cite{pietrantoni2025gsff}. Across these methods, camera pose remains the primary output and evaluation target. RIDE focuses on the render--match--PnP setting and uses the correspondences retained by PnP-RANSAC to derive sparse metric depth observations.

\subsection{Video Depth Priors}

Video-depth methods aim to produce temporally consistent depth rather than independent frame-wise estimates. Video Depth Anything (VDA) combines efficient spatiotemporal modeling with a long-video inference strategy~\cite{chen2025videodepth}, while diffusion-based methods such as DepthCrafter and ChronoDepth improve temporal consistency through video generation priors and cross-frame context~\cite{hu2025depthcrafter,shao2025chronodepth}. These models recover temporally coherent dense scene structure from video, but their output scale is not constrained by the metric geometry of a particular robot map. RIDE uses the Small variant of VDA (VDA-S) as a frozen relative-depth prior and applies global-scale, spatial, and temporal calibration using metric observations supplied by relocalization at deployment.

\subsection{Sparse-Guided and Temporal Depth Completion}

RGB-guided depth completion densifies sparse metric measurements, commonly obtained from LiDAR or reconstructed visual-inertial geometry~\cite{ma2018sparsetodense,wong2020vio}. Later approaches improved spatial propagation and long-range interaction through non-local affinities and hybrid convolution--Transformer blocks~\cite{park2020nlspn,zhang2023completionformer}. More recent methods accommodate varying input patterns or adapt depth foundation models to sparse metric guidance: DepthPrompt introduces sparse depth prompting for sensor-agnostic depth estimation~\cite{park2024depthprompting}, PriorDA combines incomplete metric priors with complete relative-depth predictions~\cite{wang2025depthprior}, and Any2Full learns pattern-agnostic scale prompts for one-stage completion~\cite{zhou2026any2full}. CAPA formulates depth completion as parameter-efficient test-time adaptation of a frozen depth model~\cite{ke2026capa}. These recent methods demonstrate the value of combining pretrained depth priors with sparse metric guidance. RIDE considers a different source of guidance: metric observations obtained from correspondences retained during 3DGS relocalization. It further evaluates the transfer of a calibrator trained with observations sampled from RGB-D videos to this deployment interface.

A smaller body of work addresses temporal depth completion. RayFusion accumulates sequential cost-volume evidence across viewpoints~\cite{kim2024rayfusion}, and SVDC fuses multiple RGB--sparse-dToF frames with a temporal consistency objective~\cite{zhu2025svdc}. Both methods assume an externally provided sparse depth map, with SVDC specifically targeting sparse dToF measurements. RIDE instead derives metric depth observations from correspondences established by image matching and retained by PnP-RANSAC during relocalization. The resulting observations are sparse, nonuniformly distributed, and can be temporarily unavailable. RIDE integrates this intermittent metric evidence through Cauchy-weighted robust scale estimation, spatial calibration, and flow-guided causal memory, without requiring additional sensors at deployment.

\section{Methodology}
\label{sec:method}

\begin{figure*}[t]
    \centering
    \includegraphics[width=0.80\textwidth]{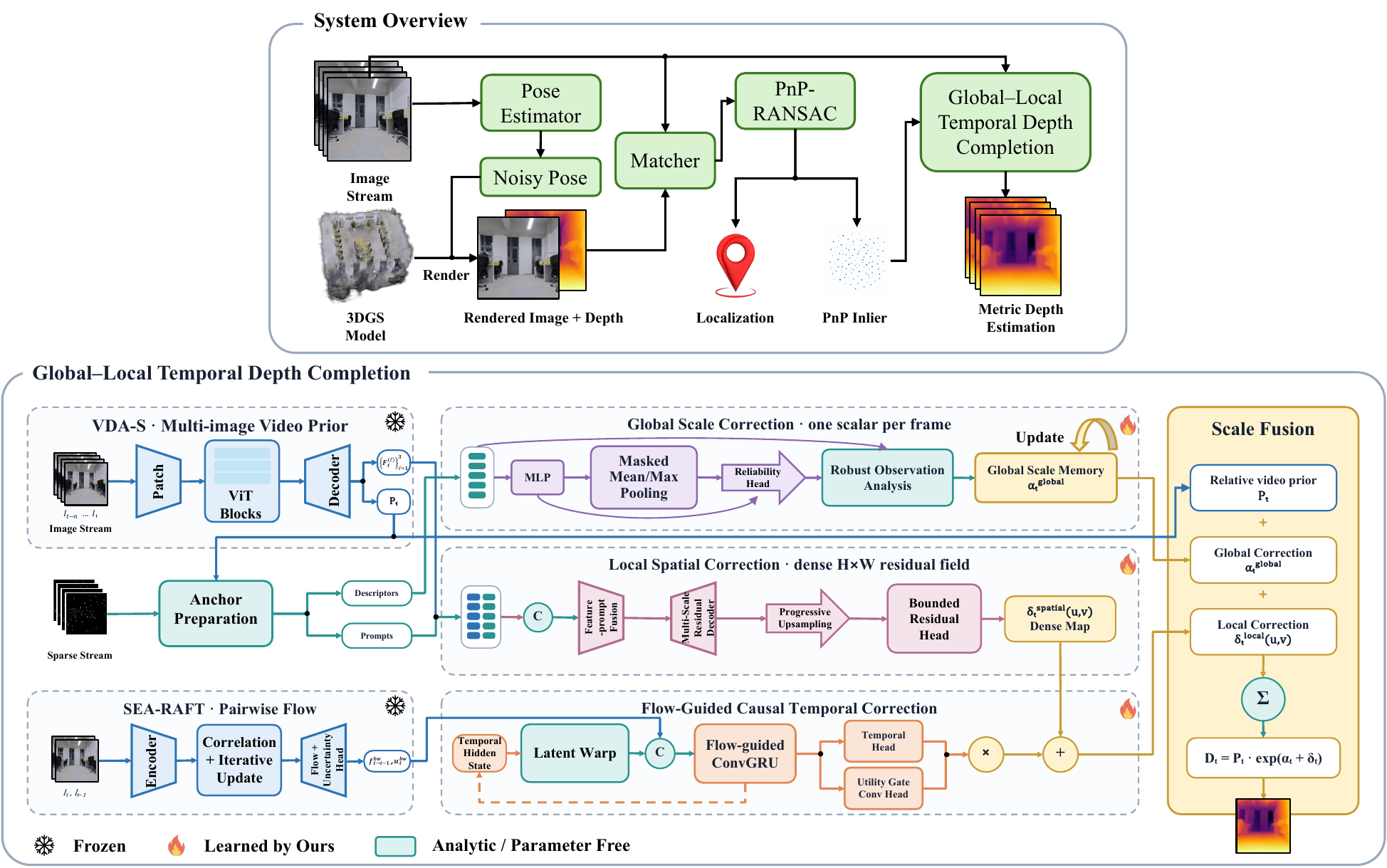}
    \caption{RIDE overview. Learned anchor reliability supports robust global scale estimation and sparse spatial prompting. Recurrent global memory and bounded spatial and temporal log-scale residuals calibrate the frozen VDA-S prior; auxiliary context inputs are omitted for clarity.}
    \label{fig:ride_overview}
\end{figure*}

Fig.~\ref{fig:ride_overview} summarizes RIDE. Sparse metric depth observations derived from PnP-RANSAC inlier correspondences complement the dense relative-depth prior and features supplied by frozen VDA-S. Learned anchor reliability guides global scale estimation and local spatial correction, while recurrent global memory and flow-guided temporal correction carry calibration information across frames. The resulting global and local corrections transform the prior into dense metric depth.

\subsection{Metric Anchors and Video Depth Prior}
\label{subsec:relocalization}

Given RGB frames up to $I_t$, a metric 3DGS model $\mathcal{G}$, camera intrinsics $\mathbf{K}$, and a coarse pose $\bar{\mathbf{T}}_{cw,t}$, RIDE estimates dense metric depth $D_t$.

The coarse pose can come from existing localization methods (Sec.~\ref{sec:related_work}); our focus is the subsequent depth estimation. Using this pose, we render virtual RGB $I_t^r$ and metric depth $Z_t^r$ from $\mathcal{G}$. For a query--render match $(\mathbf{u}_{t,i},\tilde{\mathbf{u}}_{t,i})$, valid rendered depth lifts the rendered pixel to a metric map point,
\begin{equation}
    {}^w\mathbf{X}_{t,i}
    =\bar{\mathbf{R}}_{wc,t}
      \left[Z_t^r(\tilde{\mathbf{u}}_{t,i})
      \mathbf{K}^{-1}\tilde{\mathbf{u}}_{t,i}^{h}\right]
      +\bar{\mathbf{t}}_{wc,t}.
    \label{eq:backproject}
\end{equation}
PnP-RANSAC refines the pose to $\hat{\mathbf{T}}_{cw,t}$ using the 2D to 3D correspondences $\{(\mathbf{u}_{t,i},{}^w\mathbf{X}_{t,i})\}$ and retains an inlier index set $\mathcal{J}_t$. Each retained map point yields a depth observation at its matched query pixel,
\begin{equation}
    z_{t,i}=\mathbf{e}_3^{\mathsf T}
    \left(\hat{\mathbf{R}}_{cw,t}{}^w\mathbf{X}_{t,i}
    +\hat{\mathbf{t}}_{cw,t}\right),
    \qquad \mathbf{e}_3=[0,0,1]^{\mathsf T},
    \label{eq:optical_axis_depth}
\end{equation}
giving the sparse metric observations passed to the calibrator:
\begin{equation}
    \mathcal{A}_t
    =\{(\mathbf{u}_{t,i},z_{t,i})\}_{i\in\mathcal{J}_t},
    \qquad |\mathcal{A}_t|\leq256.
    \label{eq:anchor_set}
\end{equation}

Frozen VDA-S~\cite{chen2025videodepth} supplies dense scene structure to complement these sparse metric observations. Let $R_t$ denote its raw output. From its finite positive values, we compute the per-frame floor $\zeta_t=Q_{0.001}(R_t)$ and define
\begin{equation}
\begin{aligned}
    v_t(\mathbf{u})
      &=\mathbb{I}\!\left[\operatorname{finite}(R_t(\mathbf{u}))
        \land R_t(\mathbf{u})>\zeta_t\right],\\
    P_t(\mathbf{u})
      &=\left[\operatorname{where}\!\left(
        v_t(\mathbf{u}),R_t(\mathbf{u}),\zeta_t\right)\right]^{-1}.
\end{aligned}
    \label{eq:vda_prior}
\end{equation}
This yields a positive prior $P_t$ and a floor mask $M_t^{\mathrm{floor}}=1-v_t$ marking replaced pixels. VDA-S also provides multiscale features $\{\mathbf{F}_t^{(\ell)}\}_{\ell=1}^{3}$.

At each anchor, we sample the prior depth $p_{t,i}$. Scale estimation uses nonpadded anchors with finite positive metric and prior depths outside the floor mask. Their log-scale residuals are
\begin{equation}
    r_{t,i}=\log z_{t,i}-\log p_{t,i}.
    \label{eq:anchor_residual}
\end{equation}
A shared MLP encodes each anchor's normalized coordinates, log depths, residual statistics, neighborhood support, and validity information. Masked mean and maximum pooling provide set context for a per-anchor head that predicts reliability $\eta_{t,i}\in[0,1]$, without ordering-dependent operations.

\subsection{Spatial and Temporal Metric Calibration}
\label{subsec:calibration}

We first estimate a robust global scale from the encoded anchors and update it over time. For anchors valid for scale estimation, let
\begin{equation}
    \bar r_t=\operatorname{median}_i r_{t,i},\qquad
    \gamma_t=\operatorname{median}_i|r_{t,i}-\bar r_t|+10^{-6},
\end{equation}
and define the Cauchy and combined weights
\begin{equation}
    \kappa_{t,i}=\left[1+\left(\frac{r_{t,i}-\bar r_t}{\gamma_t}\right)^2\right]^{-1},
    \qquad w_{t,i}=\eta_{t,i}\kappa_{t,i}.
    \label{eq:robust_weight}
\end{equation}
The instantaneous log-scale observation is the normalized weighted mean of the anchor residuals,
\begin{equation}
    \alpha_t^{\mathrm{obs}}
    =\frac{\sum_i w_{t,i}r_{t,i}}{\sum_i w_{t,i}}.
    \label{eq:robust_scale_observation}
\end{equation}
It is valid when at least eight scale-valid anchors are present and the denominator is finite and positive.

Let $\mathbf{g}_t=\operatorname{GAP}(\mathbf{F}_t^{(1)})$, where GAP denotes global average pooling. The Global Scale Memory predicts a small drift from the current and previous pooled features,
\begin{equation}
\begin{aligned}
    \Delta\alpha_t^{\mathrm{drift}}
      &=0.05\tanh d_\theta\!\left(
        [\mathbf{g}_t,\mathbf{g}_{t-1}]\right),\\
    \alpha_t^{-}
      &=\alpha_{t-1}^{\mathrm{global}}
        +\Delta\alpha_t^{\mathrm{drift}}.
\end{aligned}
    \label{eq:global_prediction}
\end{equation}
A gain network predicts $K_t\in[0,1]$ from anchor support and reliability, residual dispersion, prior statistics, flow confidence (Eq.~\eqref{eq:flow_confidence}), and the innovation $\alpha_t^{\mathrm{obs}}-\alpha_t^-$. After initialization, each valid observation updates the global log scale as
\begin{equation}
    \alpha_t^{\mathrm{global}}
    =\alpha_t^-+K_t(\alpha_t^{\mathrm{obs}}-\alpha_t^-).
    \label{eq:global_memory}
\end{equation}
The first valid observation initializes the state directly. If the observation is invalid, the gain is set to zero and the predicted state is retained. Metric output remains invalid until the scale is initialized.

To correct spatially varying errors beyond the global scale, we rasterize the anchors into prompts encoding weighted residuals, reliability, occupancy, neighborhood support, and the prior-floor mask. A compact top-down decoder combines these prompts with the log prior and multiscale VDA features to produce a full-resolution spatial log-scale residual,
\begin{equation}
    \delta_t^{\mathrm{sp}}(\mathbf{u})
    =0.3\tanh q_t^{\mathrm{sp}}(\mathbf{u}).
    \label{eq:spatial_delta}
\end{equation}
Here $q_t^{\mathrm{sp}}$ is the decoder's raw spatial logit. The bound limits the branch to local multiplicative calibration while the frozen prior supplies the underlying scene structure.

To reuse local calibration information across frames, we align the earlier latent state using optical flow. Frozen SEA-RAFT~\cite{wang2024searaft} is evaluated in both directions,
\begin{equation}
    \mathbf{b}_t=\operatorname{RAFT}(I_t,I_{t-1}),\qquad
    \mathbf{f}_t=\operatorname{RAFT}(I_{t-1},I_t),
\end{equation}
where only the current-to-previous flow $\mathbf{b}_t$ enters the temporal network. The operator $\mathcal{B}$ samples a map at the pixel coordinates specified by its second argument. Let $\widehat{\mathbf{f}}_t(\mathbf{u})=\mathcal{B}(\mathbf{f}_t,\mathbf{u}+\mathbf{b}_t(\mathbf{u}))$ and define the forward--backward error as $e_t^{fb}=\|\mathbf{b}_t+\widehat{\mathbf{f}}_t\|_2$. With SEA uncertainty $U_t$ and in-bounds mask $m_t^b$, the flow confidence is
\begin{equation}
    c_t=m_t^b\exp[-(e_t^{fb}/3)^2]\exp[-U_t/5].
    \label{eq:flow_confidence}
\end{equation}
The earlier latent state is aligned to the current frame by
\begin{equation}
    \widetilde{\mathbf{H}}_{t-1}(\mathbf{u})
    =c_t(\mathbf{u})m_t^b(\mathbf{u})
      \mathcal{B}(\mathbf{H}_{t-1},\mathbf{u}+\mathbf{b}_t(\mathbf{u})).
    \label{eq:latent_warp}
\end{equation}
A ConvGRU combines the aligned history with current VDA features, prior and anchor context, global scale, spatial logit $q_t^{\mathrm{sp}}$, and flow validity cues. Temporal and utility heads predict proposal logits $q_t^{\mathrm{tmp}}$ and a gate $g_t$. With $\mathcal{U}_{\mathrm{guided}}$ denoting guided upsampling to the resolution of $P_t$, the temporal correction is
\begin{equation}
\begin{aligned}
    \delta_t^{\mathrm{tmp}}
      &=\mathcal{U}_{\mathrm{guided}}\!\left[
        g_t\,0.3\tanh q_t^{\mathrm{tmp}}\right],\\
    g_t&\in[0,1],\qquad
      |\delta_t^{\mathrm{tmp}}|<0.3.
\end{aligned}
    \label{eq:temporal_delta}
\end{equation}
Missing flow zeros the warped history while preserving global memory.

The local correction and final metric depth are
\begin{equation}
\begin{aligned}
    \delta_t^{\mathrm{local}}
      &=\delta_t^{\mathrm{sp}}+\delta_t^{\mathrm{tmp}},\\
    D_t(\mathbf{u})
      &=P_t(\mathbf{u})\exp\!\left(
        \alpha_t^{\mathrm{global}}
        +\delta_t^{\mathrm{local}}(\mathbf{u})\right).
\end{aligned}
    \label{eq:final_depth}
\end{equation}
Inference is causal: $D_t$ uses only images and anchors available up to time $t$ and recurrent states from earlier inputs.

\subsection{Staged Training and Supervision}
\label{subsec:training}

Common depth-completion benchmarks use LiDAR scan lines or subsampled structured-light depth~\cite{ma2018sparsetodense,zhang2023completionformer}. These spatial patterns differ from the irregular feature locations obtained by image matching. Monocular structure-from-motion (SfM) yields sparse geometry with no inherent absolute scale unless externally aligned~\cite{zuo2025omnidc}. To our knowledge, no large-scale public dataset jointly provides the 2D to 3D correspondences retained by PnP-RANSAC during 3DGS relocalization and dense metric video depth. We therefore sample registered metric depth at detected feature locations in public RGB-D videos~\cite{tan2026masked} to emulate the deployment anchor interface. Training uses these RGB-D-derived observations, whereas deployment uses observations derived from PnP-RANSAC inlier correspondences; both are represented as pixel--depth pairs in $\mathcal{A}_t$ (Eq.~\eqref{eq:anchor_set}).

Let $D_t^*$ denote the registered metric reference depth. The depth losses $\mathcal{L}_{\mathrm{hold}}$ and $\mathcal{L}_{\mathrm{dense}}$ supervise predictions at clean held-out anchor locations and valid dense reference pixels, respectively. The frame-wise depth, global-scale, and proposal regression terms use Smooth-L1. Scalar loss weights $w_1,\ldots,w_{15}$ and loss hyperparameters are specified in Sec.~\ref{subsec:experimental_setup}.

Stage~1 trains the reliability module using focal binary cross-entropy and supervision from clean held-out anchors,
\begin{equation}
    \mathcal{L}_1=w_1\mathcal{L}_{\mathrm{gate}}+w_2\mathcal{L}_{\mathrm{hold}}.
    \label{eq:stage1_loss}
\end{equation}
Stage~2 freezes VDA-S and the reliability module and uses $\alpha_t^{\mathrm{obs}}$ directly as the global log scale. The spatial decoder is trained using
\begin{equation}
    \mathcal{L}_2=w_2\mathcal{L}_{\mathrm{hold}}
    +w_3\mathcal{L}_{\mathrm{dense}}
    +w_4\mathcal{L}_{\mathrm{excl}}
    +w_5\mathcal{L}_{\mathrm{smooth}}
    +w_6\mathcal{L}_{\mathrm{reg}},
    \label{eq:stage2_loss}
\end{equation}
where $\mathcal{L}_{\mathrm{excl}}$ is an auxiliary dense-depth training loss evaluated outside a four-pixel neighborhood of each input anchor, $\mathcal{L}_{\mathrm{smooth}}$ is edge-aware, and $\mathcal{L}_{\mathrm{reg}}$ penalizes the spatial residual magnitude. The depth terms are evaluated only on frames with a valid scale observation.

Stage~3 introduces the global and temporal recurrent modules while keeping VDA-S, SEA-RAFT, the reliability module, and the spatial decoder frozen. These modules are trained on five-frame chunks using truncated backpropagation through time (TBPTT). For frame $t$, let
\begin{equation}
    \mathcal{S}_t=w_2\mathcal{L}_{\mathrm{hold},t}
    +w_3\mathcal{L}_{\mathrm{dense},t}
    +w_4\mathcal{L}_{\mathrm{excl},t}.
\end{equation}
Let $\overline{\mathcal{S}}_{1:4}=\frac{1}{4}\sum_{t=1}^{4}\mathcal{S}_t$. The Stage~3 objective is
\begin{align}
    \mathcal{L}_3={}&w_7\mathcal{S}_5
    +w_8\overline{\mathcal{S}}_{1:4}
    +w_9\mathcal{L}_{\mathrm{TGM}}
    +w_{10}\mathcal{L}_{\alpha}\nonumber\\
    &+w_{11}\mathcal{L}_{\mathrm{proposal}}
    +w_{12}\mathcal{L}_{\mathrm{utility}}\nonumber\\
    &+w_{13}\operatorname{mean}_{t,\mathbf{u}}
      |\delta_t^{\mathrm{tmp}}(\mathbf{u})|\nonumber\\
    &+w_{14}\mathcal{L}_{\mathrm{localmean}}
      +w_{15}\operatorname{mean}_t
      |\Delta\alpha_t^{\mathrm{drift}}|.
    \label{eq:stage3_loss}
\end{align}
The loss $\mathcal{L}_{\mathrm{TGM}}$ aligns log-depth errors across flow correspondences, and $\mathcal{L}_{\alpha}$ supervises global scale using the median dense log-scale target. The proposal target for $\delta_t^{\mathrm{tmp}}$ is the reference log depth minus the prediction after global and spatial correction, clipped to $[-0.3,0.3]$. The utility loss supervises the gate with a soft target based on the local reduction in absolute log-depth error from temporal correction, where depth validity, flow confidence, and scale initialization are sufficient. Both targets are detached from gradient computation. The loss $\mathcal{L}_{\mathrm{localmean}}$ penalizes the squared spatial mean of $\delta_t^{\mathrm{tmp}}$. Held-out anchors, dense depth, outlier labels, and utility targets serve only as supervision or evaluation references.

\section{Experiments}
\label{sec:experiments}

\subsection{Experimental Setup}
\label{subsec:experimental_setup}

\paragraph{Public RGB-D data.}
We use the RobbyReal subset of LingBot-Depth~\cite{tan2026masked}, which provides RGB video, registered metric depth, and camera intrinsics. Its 183 indoor scenes are partitioned into 147/18/18 training/validation/test scenes, with all camera sequences from a scene assigned to the same split. After excluding invalid calibrations and target frames without valid depth, the splits contain 134,787/14,815/15,451 causal clips, respectively. Each clip contains five consecutive frames ending at the target frame, on which frame-wise metrics are evaluated. Registered depths between 0.3 and 20\,m are valid. Model selection and hyperparameter choices use only the validation split.

We sample valid registered depth at detected feature locations to form RGB-D-derived observations. During training, approximately $20\%$ are held out for supervision before each forward pass. The input observations undergo dropout at a rate sampled from $[0.2,0.8]$ and are capped at a count selected uniformly from $\{32,64,128,256\}$. We then introduce $0$--$30\%$ structured outliers to simulate background leakage, spatially clustered errors, or incorrect depth associations. Input and held-out observations remain disjoint. Stage~3 uses a fixed mixture of clean, anchor-outage, prior-flicker, persistent-outlier, and flow-failure chunks.

\paragraph{Robot sequences.}
We additionally evaluate 3,012 frames from 27 routes in five real-world scenes. Each route is processed continuously, with the global scale and temporal hidden states reset only at route boundaries. Sparse metric inputs are derived from PnP-RANSAC inlier correspondences against a metric 3DGS model (Sec.~\ref{subsec:relocalization}). We use the weights trained on public RGB-D videos without robot-data training or fine tuning. Registered RGB-D measurements serve only as evaluation references and are not supplied to the relocalization front end or depth model.

\paragraph{Metrics and evaluation support.}
For valid evaluation pixels $\Omega_t$, we report
\begin{equation}
    \begin{aligned}
    \operatorname{AbsRel}_t
      &=\frac{1}{|\Omega_t|}
        \sum_{\mathbf{u}\in\Omega_t}
        \frac{|D_t(\mathbf{u})-D_t^*(\mathbf{u})|}{D_t^*(\mathbf{u})},\\
    \delta_{1,t}
      &=\frac{1}{|\Omega_t|}
        \sum_{\mathbf{u}\in\Omega_t}
        \mathbb{I}\!\left[
        \max\!\left(
        \frac{D_t(\mathbf{u})}{D_t^*(\mathbf{u})},
        \frac{D_t^*(\mathbf{u})}{D_t(\mathbf{u})}
        \right)<1.25
        \right].
    \end{aligned}
    \label{eq:metrics}
\end{equation}
Held-out error is the mean absolute log-depth error at clean observations withheld from the model input. We report it only on public RGB-D data; robot observations are deployment inputs and are not partitioned into input and held-out sets. To measure temporal accuracy, let

$\Delta_t(D,\mathbf{u})=\log D_t(\mathbf{u})-
\log\mathcal{B}(D_{t-1},\mathbf{u}+\mathbf{b}_t(\mathbf{u}))$.
Temporal geometric error is
\begin{equation}
    \operatorname{TGE}_t
    =\frac{1}{|\Omega_t^{\mathrm{temp}}|}
      \sum_{\mathbf{u}\in\Omega_t^{\mathrm{temp}}}
      \left|\Delta_t(D,\mathbf{u})-\Delta_t(D^*,\mathbf{u})\right|,
    \label{eq:tge}
\end{equation}
where $\Omega_t^{\mathrm{temp}}$ contains correspondences satisfying reference-depth validity in both frames, in-bounds flow, and the common temporal validity mask. We average $\operatorname{TGE}_t$ over eligible consecutive target-frame pairs. All methods share the precomputed flow and reference-validity masks. On the robot sequences, all methods process the same 3,012 frames, but each is scored on the intersection of its own valid-output mask and the required reference mask.

\paragraph{Baselines.}
We include three calibration baselines built on frozen VDA-S: uniform median estimates a current-frame scale from input anchors; robust scale combines learned reliability with Cauchy weights to average log-scale residuals; and the spatial variant adds the local decoder without temporal correction. Any2Full~\cite{zhou2026any2full}, PriorDA v1.1~\cite{wang2025depthprior}, and DepthPrompt~\cite{park2024depthprompting} receive the same sparse input anchors under their respective inference protocols. For DA3Mono-Large~\cite{lin2025depthanything3}, oVDA-S c8~\cite{feiden2025onlinevda}, and Marigold v1-0~\cite{ke2024marigold}, ``+ LS'' denotes a scale $s$ fitted to a relative prediction $Q_t$ using only current-frame input anchors:
$\min_{s>0}\sum_i\left(sQ_t(\mathbf{u}_i)-z_i\right)^2$.
No baseline uses held-out anchors or dense ground truth for calibration.

\paragraph{Implementation.}
Stages~1 and~2 construct the 32-frame VDA-S input from five distinct frames followed by 27 repetitions of the target frame and supervise only the target. They use AdamW with learning rate $10^{-3}$ for 20 and 5 epochs, respectively. Stage~3 uses stateful causal VDA-S streaming at $518\times924$, four GPUs, and five-frame TBPTT. It supervises all five frames, emphasizing the last; recurrent states are carried between contiguous chunks with gradients detached at chunk boundaries. VDA-S and SEA-RAFT remain frozen throughout. We use mixed precision and perform no test-time parameter updates.

\paragraph{Loss weights.}
For the objectives in Sec.~\ref{subsec:training}, we use $(w_1,\ldots,w_6)=(1,1,0.2,0.05,0.02,0.01)$, $(w_7,w_8)=(1,0.2)$, and $(w_{10},\ldots,w_{15})=(0.5,0.25,0.1,0.005,0.01,0.01)$. The temporal weight $w_9$ is selected on the validation split and then fixed for all experiments. Smooth-L1 uses $\beta=1$, and focal binary cross-entropy uses $\gamma=2$.

\subsection{Relocalization-Derived Metric Anchors}
\label{subsec:anchor_analysis}

\begin{figure*}[t]
    \centering
    \includegraphics[width=0.8\textwidth]{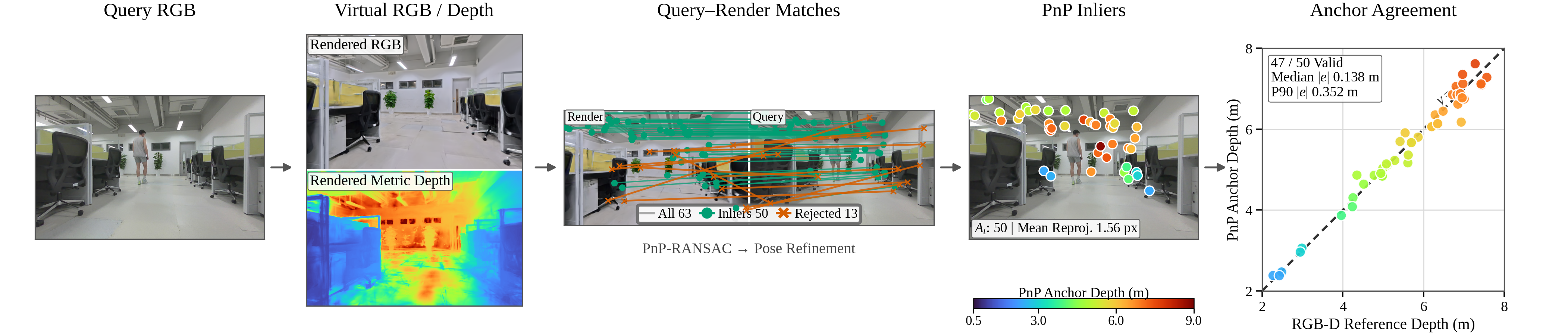}
    \caption{Metric anchor extraction from query--render correspondences and agreement with registered RGB-D reference depth in one robot frame.}
    \label{fig:relocalization_anchors}
\end{figure*}

Fig.~\ref{fig:relocalization_anchors} traces the workflow from relocalization to metric anchor extraction. A coarse pose is used to render RGB and depth from the metric 3DGS model. Matched rendered pixels are lifted to map points using rendered depth and paired with query pixels to form 2D to 3D correspondences for PnP-RANSAC. After inlier selection and pose refinement, the retained map points are transformed into the query camera frame to obtain optical-axis depth observations at the matched query pixels.

The example contains a person absent from the map. PnP-RANSAC retains 50 of 63 correspondences, with a mean inlier reprojection error of 1.56 pixels. Of the resulting anchors, 47 have valid RGB-D references; their median and 90th-percentile absolute depth differences are 0.138\,m and 0.352\,m. Reprojection consistency and agreement with reference depth provide complementary checks of the anchor-extraction workflow in this frame.

\subsection{Comparison with Baselines}
\label{subsec:quantitative_comparison}

\begin{figure}[t]
    \centering
    \includegraphics[width=0.9\columnwidth]{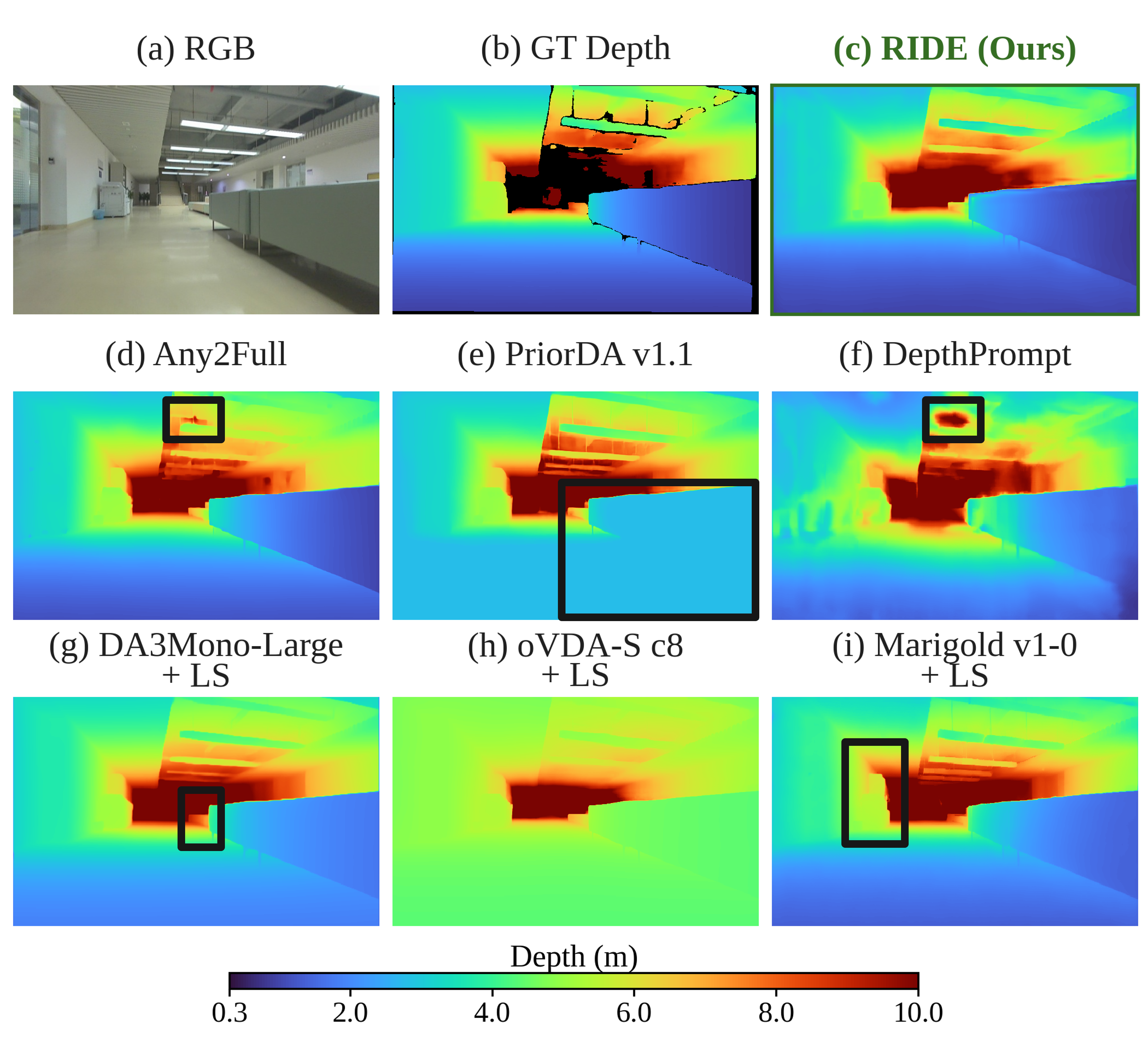}
    \caption{Qualitative comparison on a real-world robot scene. All depth maps use the same metric color scale; boxes mark representative local artifacts.}
    \label{fig:robot_qualitative_comparison}
\end{figure}

Table~\ref{tab:baseline_comparison} compares RIDE with the calibration, sparse-guided completion, and foundation-depth baselines on public RGB-D data and robot sequences.

\begin{table*}[t]
    \centering
    \caption{Comparison on LingBot and robot sequences (five scenes, 27 routes). Lower is better except for $\delta_1$. Best results are bold; second-best results are underlined.}
    \label{tab:baseline_comparison}
    \setlength{\tabcolsep}{3pt}
    \footnotesize
    \begin{tabular}{@{}l*{4}{>{\centering\arraybackslash}m{1.50cm}}@{\hspace{8pt}}*{3}{>{\centering\arraybackslash}m{1.50cm}}@{}}
        \toprule
        & \multicolumn{4}{c}{LingBot} & \multicolumn{3}{c}{Real robot} \\
        \cmidrule(lr){2-5}\cmidrule(lr){6-8}
        Method & AbsRel $\downarrow$ & $\delta_1$ $\uparrow$ & Held-out $\downarrow$ & TGE $\downarrow$
        & AbsRel $\downarrow$ & $\delta_1$ $\uparrow$ & TGE $\downarrow$ \\
        \midrule
        VDA-S + uniform median
            & 0.2916 & 0.7226 & 0.1571 & 0.0857 & 0.2013 & 0.5852 & 0.0533 \\
        VDA-S + robust scale
            & 0.2905 & 0.7228 & 0.1568 & 0.0842 & 0.2078 & 0.5646 & \underline{0.0530} \\
        VDA-S + robust scale + spatial decoder
            & 0.1759 & 0.8994 & 0.0800 & \underline{0.0674} & \underline{0.1056} & \underline{0.8919} & 0.0538 \\
        Any2Full
            & 0.1617 & 0.9136 & \underline{0.0722} & 0.0963 & 0.2525 & 0.6616 & 0.0849 \\
        PriorDA v1.1
            & \textbf{0.0917} & \underline{0.9175} & 0.0764 & 0.0783 & 0.4188 & 0.6419 & 0.1041 \\
        DepthPrompt
            & 0.3124 & 0.6360 & 0.1443 & 0.1828 & 0.7086 & 0.3617 & 0.1625 \\
        DA3Mono-Large + LS
            & 0.2202 & 0.7823 & 0.1426 & 0.1570 & 0.2605 & 0.6852 & 0.1949 \\
        oVDA-S c8 + LS
            & 0.2968 & 0.6845 & 0.1814 & 0.1546 & 0.7993 & 0.4206 & 0.0944 \\
        Marigold v1-0 + LS
            & 0.2546 & 0.7269 & 0.1643 & 0.1599 & 0.3476 & 0.6423 & 0.1918 \\
        \textbf{RIDE (Ours)}
            & \underline{0.1115} & \textbf{0.9322} & \textbf{0.0675} & \textbf{0.0394}
            & \textbf{0.0927} & \textbf{0.9361} & \textbf{0.0298} \\
        \bottomrule
    \end{tabular}
\end{table*}

On public RGB-D data, RIDE achieves the highest $\delta_1$ and the lowest held-out error and TGE, while PriorDA v1.1 obtains the lowest AbsRel. The higher threshold accuracy indicates that more evaluated pixels meet the relative-error criterion, and the lower held-out error shows accurate calibration at feature locations not supplied as input anchors. RIDE also reduces TGE from the best competing value of 0.0674, obtained by the VDA-S spatial variant, to 0.0394 (41.5\%). This reduction indicates closer agreement with reference temporal depth changes, complementing the frame-wise accuracy results.

On robot sequences, RIDE achieves the best reported values on all three metrics under the stated valid-output protocol. Compared with the VDA-S spatial variant, AbsRel decreases from 0.1056 to 0.0927 and $\delta_1$ increases from 0.8919 to 0.9361, showing gains in both average relative error and threshold accuracy. TGE falls from the best competing value of 0.0530 to 0.0298 (43.8\%). These results support transfer of the learned calibration from RGB-D-derived training observations to relocalization-derived inputs without robot-data fine tuning, with improvements in both frame-wise and temporal accuracy.

Fig.~\ref{fig:robot_qualitative_comparison} provides a visual comparison in a robot scene. RIDE recovers coherent lobby geometry and clear structural boundaries; the marked regions highlight local artifacts or depth discrepancies in several baseline predictions.

\subsection{Ablation Studies}
\label{subsec:ablations}

\begin{table}[H]
    \centering
    \caption{Ablation results on the LingBot validation split. Best results are bold; second-best results are underlined.}
    \label{tab:ablation}
    \footnotesize
    \setlength{\tabcolsep}{3pt}
    \begin{tabular}{@{}lcccc@{}}
        \toprule
        Method & AbsRel $\downarrow$ & $\delta_1$ $\uparrow$ & Held-out $\downarrow$ & TGE $\downarrow$ \\
        \midrule
        Global only
            & 0.2173 & 0.7450 & 0.1515 & 0.0713 \\
        Spatial only
            & 0.1178 & 0.9126 & 0.0676 & 0.0513 \\
        \midrule
        No state carryover
            & \underline{0.0967} & \underline{0.9338} & \underline{0.0588} & 0.0438 \\
        Shuffled history
            & 0.1201 & 0.9057 & 0.0856 & 0.0754 \\
        Shuffled flow
            & 0.0986 & 0.9329 & 0.0621 & 0.0492 \\
        No flow-confidence mask
            & 0.1052 & 0.9268 & 0.0693 & \underline{0.0370} \\
        \textbf{Full RIDE}
            & \textbf{0.0947} & \textbf{0.9366} & \textbf{0.0576} & \textbf{0.0354} \\
        \bottomrule
    \end{tabular}
\end{table}

All ablations use the LingBot validation split with identical clips, anchor realizations, and evaluation masks. Global only uses the instantaneous Cauchy-robust scale observation; Spatial only adds the local decoder. Neither includes the global scale memory or temporal adapter. We then compare Full RIDE with inference-time interventions on the same checkpoint: resetting recurrent state at every frame, shuffling history, shuffling flow, or setting $c_t\equiv1$ to remove flow-confidence weighting.

Table~\ref{tab:ablation} evaluates the corrections progressively. Global only provides a metric scale estimate, while adding the spatial decoder improves all four metrics. AbsRel decreases from 0.2173 to 0.1178, and held-out error from 0.1515 to 0.0676. These gains support local correction beyond a single global scale and extend to feature locations not supplied as input anchors.

Adding the global scale memory and temporal adapter jointly improves all four metrics, with TGE decreasing from 0.0513 to 0.0354. Resetting recurrent state at every frame raises TGE to 0.0438, supporting the benefit of carrying information across frames. Shuffling history causes the largest degradation across all four metrics, indicating that the recurrent model depends on relevant historical context.

The flow interventions further examine how that context is used. Shuffling flow raises TGE from 0.0354 to 0.0492, supporting the role of correct motion alignment. Removing flow-confidence weighting degrades all four metrics, including an increase in held-out error from 0.0576 to 0.0693. Together, these results support the complementary roles of spatial refinement, recurrent memory, flow alignment, and confidence weighting in Full RIDE.

\subsection{Qualitative Temporal Analysis}
\label{subsec:qualitative_temporal}

\begin{figure*}[t]
    \centering
    \includegraphics[width=0.85\textwidth]{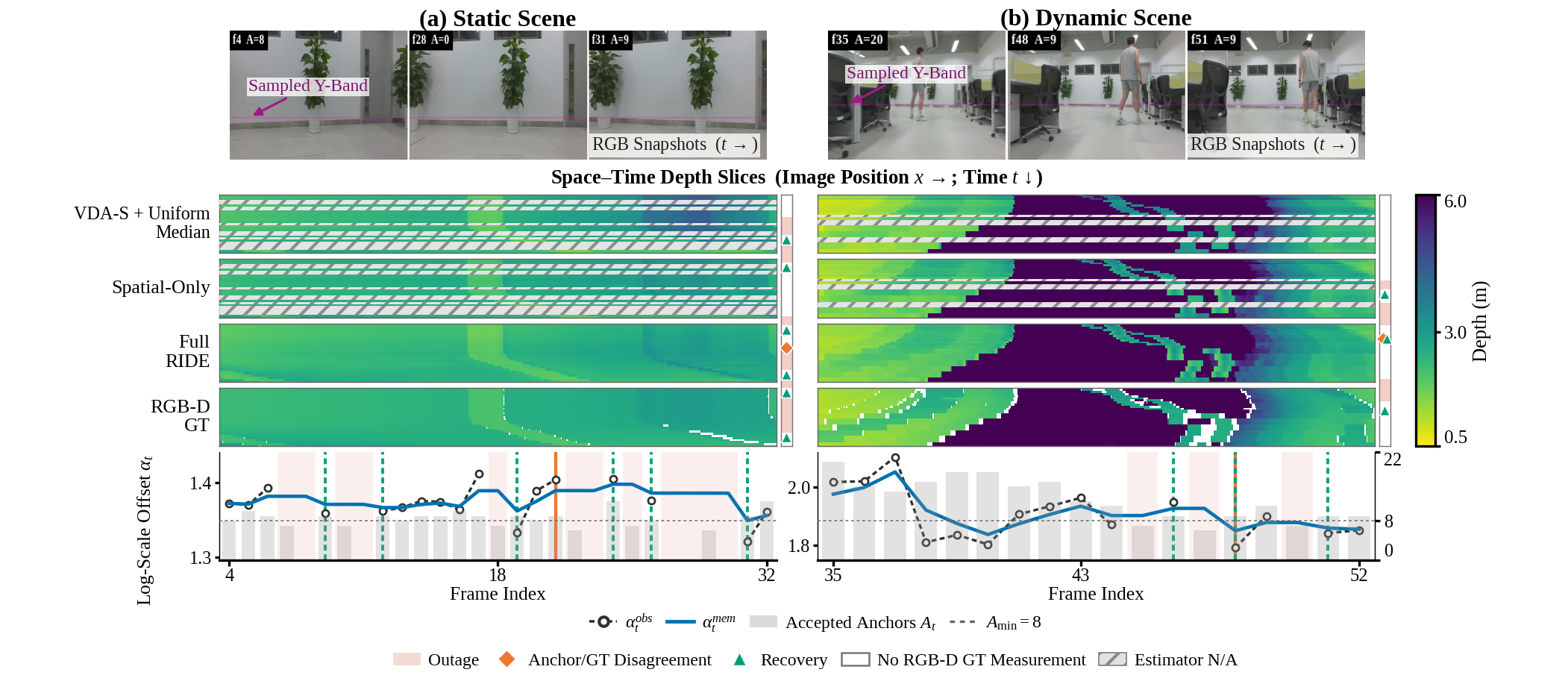}
    \caption{Temporal depth behavior under intermittent anchor outage and foreground motion. The global log-scale state $\alpha_t^{\mathrm{global}}$ is labeled $\alpha_t^{\mathrm{mem}}$ in the plots.}
    \label{fig:temporal_qualitative}
\end{figure*}

Fig.~\ref{fig:temporal_qualitative} shows space--time depth slices, with image position along the horizontal axis and time increasing downward. In the static sequence, insufficient anchor support interrupts the frame-wise estimators, producing the hatched bands of unavailable predictions. After scale initialization, RIDE maintains coherent depth through these intervals by carrying its recurrent states, reducing interruptions caused by intermittent anchor availability.

In the dynamic sequence, the foreground boundary changes position over time. RIDE preserves its trajectory, as seen by comparison with the reference depth slice, while maintaining continuous predictions through anchor shortages. This illustrates temporal continuity alongside sensitivity to foreground motion.

The lower plots relate this behavior to global scale estimation. The memory trajectory varies more smoothly than the instantaneous anchor-based observation and does not follow every fluctuation. When valid anchors return, observation-based updates resume without resetting the accumulated scale state. Together, the depth slices and scale curves illustrate continuity during short outages and responsiveness to new observations.

\section{Conclusion}

We presented RIDE, which combines camera relocalization and dense metric video-depth estimation in a single system. Given a metric 3DGS model, the relocalization front end recovers camera pose and retains PnP-RANSAC inlier correspondences. RIDE derives sparse metric depth observations from these correspondences and uses them to calibrate a frozen VDA-S prior through robust global-scale estimation, bounded spatial correction, and flow-guided temporal memory. On public RGB-D data, RIDE achieves the highest $\delta_1$ and the lowest held-out error and TGE among the evaluated methods. After training with RGB-D-derived observations, RIDE is evaluated on robot sequences using relocalization-derived observations without fine tuning. It achieves the best reported results on all three metrics across 27 routes under the stated valid-output protocol. Ablations support the benefits of spatial correction and reliable temporal context, while qualitative examples show continuous depth output through short anchor outages after scale initialization and preservation of moving foreground boundaries. These results demonstrate the value of reusing relocalization correspondences for dense robot perception. Future work will address initialization with very sparse anchors and recovery from prolonged loss of map overlap or reliable flow.

\bibliographystyle{IEEEtran}
\bibliography{references}

\end{document}